\documentclass[preprint,12pt,authoryear]{elsarticle}

\usepackage{amssymb}
\usepackage{amsmath}
\usepackage{booktabs}
\usepackage{url}
\usepackage{subcaption}
\usepackage{diagbox}

\journal{Computerized Medical Imaging and Graphics}

\begin{document}

\begin{frontmatter}



\title{Higher Chest X-ray Resolution Improves Classification Performance}


\author[label1]{Alessandro Wollek\corref{alessandro.wollek@tum.de}}
\author[label2]{Sardi Hyska}
\author[label2]{Bastian Sabel}
\author[label2]{Michael Ingrisch}
\author[label1]{Tobias Lasser}

\affiliation[label1]{organization={Munich Institute of Biomedical Engineering, and School of Computation, Information, and Technology, Technical University of Munich},
            addressline={Boltzmannstr. 11},
            city={Garching},
            postcode={85748},
            state={Bavaria},
            country={Germany}}
\affiliation[label2]{organization={Department of Radiology, University Hospital Ludwig-Maximilians-University},
            addressline={Marchioninistr. 15},
            city={Munich},
            postcode={81337},
            state={Bavaria},
            country={Germany}}

\begin{abstract}
  Deep learning models for image classification are often trained at a resolution of $224\times224$ pixels for historical and efficiency reasons.
  However, chest X-rays are acquired at a much higher resolution to display subtle pathologies.
  This study investigates the effect of training resolution on chest X-ray classification performance, using the chest X-ray 14 dataset.
  The results show that training with a higher image resolution, specifically $1024\times1024$ pixels, results in the best overall classification performance with a mean AUC of 84.2 \% compared to 82.7 \% when trained with $256\times256$ pixel images.
  Additionally, comparison of bounding boxes and GradCAM saliency maps suggest that low resolutions, such as $256\times256$ pixels, are insufficient for identifying small pathologies and force the model to use spurious discriminating features.
  Our code is publicly available at \url{https://gitlab.lrz.de/IP/cxr-resolution}.
\end{abstract}




\end{frontmatter}


\section{Introduction}
\begin{figure}
  \centering
  \includegraphics[height=0.5\textheight]{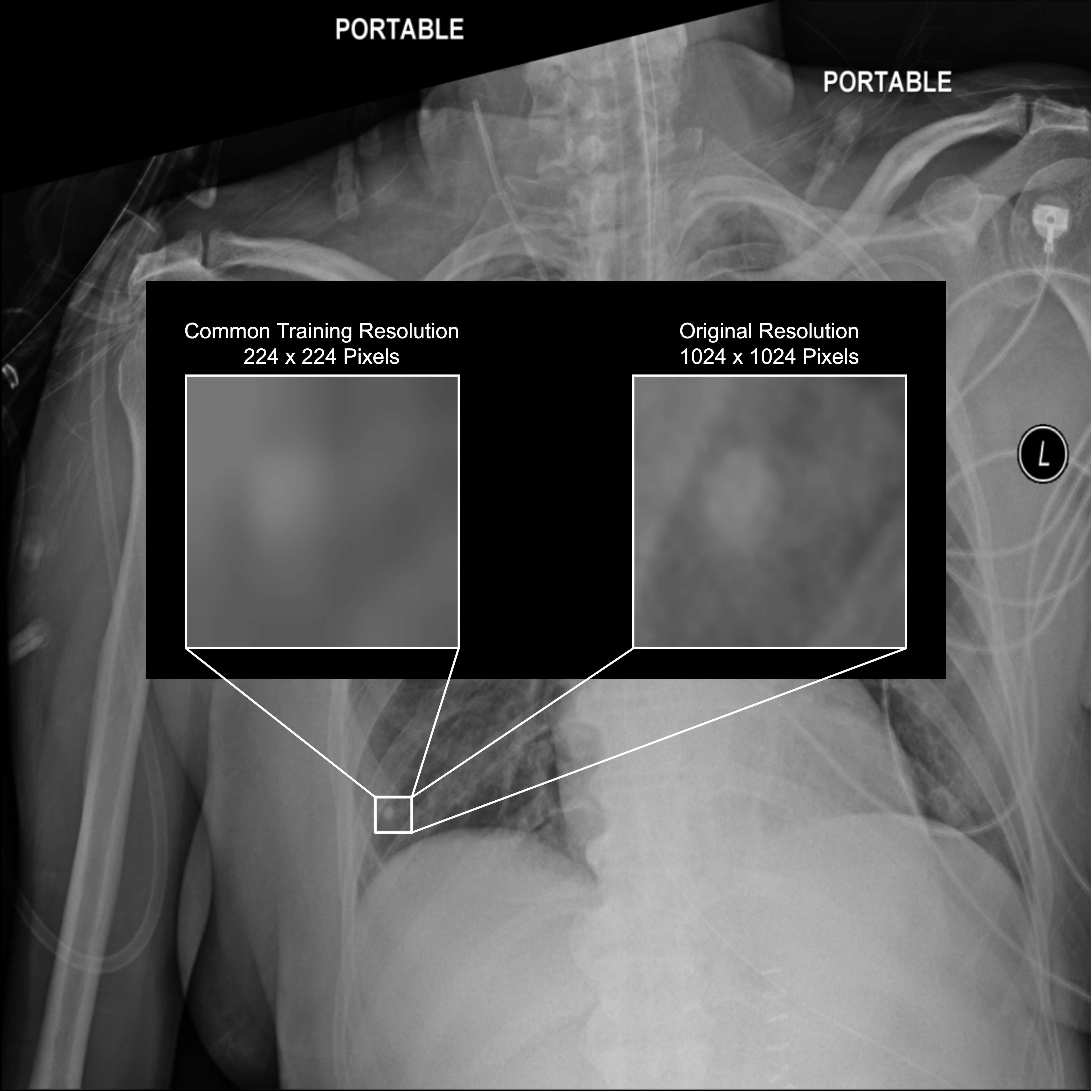}
  \caption{Chest X-ray classification models are commonly trained at a $224\times224$ pixel resolution for historical and efficiency reasons. Chest X-rays, on the other hand, are acquired with a high resolution to display subtle features. As a consequence, at a lower resolution, a small nodule (highlighted in the white bounding box) becomes blurred and almost invisible.}
  \label{fig:res:motivation}
\end{figure}

Since AlexNet, images processed by deep learning models are often resized to $224\times224$ pixels during training mostly for computational reasons \citep{krizhevskyImagenetClassificationDeep2012,heDeepResidualLearning2016,huangDenselyConnectedConvolutional2017,tanEfficientNetRethinkingModel2020,wollekAttentionbasedSaliencyMaps2023}.
Training at a lower resolution requires less memory and consequently models train faster.
However, lowering the resolution can also blur or occlude important regions of an image, as shown in Figure~\ref{fig:res:motivation}.

\cite{tanEfficientNetRethinkingModel2020} studied the importance of image resolution on image classification accuracy using the ImageNet data set~\citep{ImageNet}.
They concluded that image resolution is a hyper-parameter similar to network depth or width.
For chest radiographs, \cite{sabottkeEffectImageResolution2020a} tested different image resolutions ($32\times32$ up to $600\times600$ pixels) on the task of chest X-ray classification.
In their experiments, maximum classification AUCs were obtained between $256\times256$ and $448\times448$ pixel resolution.
While their results suggest that a $448\times448$ pixel resolution is sufficient, chest radiographs have a much higher resolution.
For example, the images in the MIMIC data set have an average resolution of $2500\times3500$ pixels~\citep{johnsonMIMICCXRDeidentifiedPublicly2019}.
To close this research gap, we investigate the effect of image resolution on chest X-ray classification performance.
Our contributions are:
\begin{itemize}
  \item We systematically analyze the effect of image resolution on chest X-ray classification performance from $64\times64$ up to $1024\times1024$ pixels, the highest resolution available.
  \item We show that training with the highest available resolution, $1024\times1024$ pixels, achieves the highest classification performance on average and for most classes (11/14).
  \item By analyzing saliency map-extracted bounding boxes, we provide evidence that training at a lower resolution encourages the network to learn spurious discriminating features, potentially eroding trust into the (correct) prediction.
\end{itemize}

\section{Materials and Methods}
\begin{table}
  \centering
\begin{tabular}{lrrr}
{Class} &    Training &  Validation &    Test \\
\midrule
Atelectasis        &   7996 &      1119 &  2420 \\
Cardiomegaly       &   1950 &       240 &   582 \\
Consolidation      &   3263 &       447 &   957 \\
Edema              &   1690 &       200 &   413 \\
Effusion           &   9261 &      1292 &  2754 \\
Emphysema          &   1799 &       208 &   509 \\
Fibrosis           &   1158 &       166 &   362 \\
Hernia             &    144 &        41 &    42 \\
Infiltration       &  13914 &      2018 &  3938 \\
Mass               &   3988 &       625 &  1133 \\
Nodule             &   4375 &       613 &  1335 \\
Pleural Thickening &   2279 &       372 &   734 \\
Pneumonia          &    978 &       133 &   242 \\
Pneumothorax       &   3705 &       504 &  1089 \\
\end{tabular}

  \caption{Class distributions of the data set used in this study.}
  \label{res:tab:dataset}
\end{table}
\begin{table}
  \centering
\begin{tabular}{lrr}
Finding       &  Bounding Box Area ($px^{2}$) & \#Samples \\
\midrule
Nodule        &    5,527 &  18 \\
Atelectasis   &   33,753 &  43 \\
Mass          &   50,083 &  20 \\
Pneumothorax  &   55,208 &  18 \\
Effusion      &   61,901 &  30 \\
Infiltration  &  119,563 &  25 \\
Pneumonia     &  159,466 &  22 \\
Cardiomegaly  &  184,134 &  28 \\
\end{tabular}
\caption{Mean bounding box area per class in squared pixels. Annotated samples stem from the provided test split.}
\label{res:tab:area}
\end{table}

\subsection{Data Set}

For our experiments we chose the publicly available chest X-ray 14 data set containing 112,120 frontal view chest radiographs from 32,717 patients~\citep{wangChestXray8HospitalscaleChest2017}.
The chest radiographs were annotated according to the 14 labels atelectasis, cardiomegaly, consolidation, edema, effusion, emphysema, fibrosis, hernia, infiltration, mass, nodule, pleural thickening, pneumonia, and pneumothorax.
We selected this data set specifically as the authors provide a small test sub set (983 images) with bounding box annotations for eight of the 14 classes (atelectasis, cardiomegaly, effusion, infiltration, mass, nodule, pneumonia, and pneumothorax).
For training, we used the data split provided by the authors.
The class distributions are reported in Table~\ref{res:tab:dataset}.
While the images of the chest X-ray 14 data set were down-scaled to $1024\times1024$ pixels before their release \citep{wangChestXray8HospitalscaleChest2017}, it is the only large, publicly available data set that also contains bounding boxes (see Table~\ref{res:tab:area}).

\subsection{Chest X-ray Classification}
For classification, we used the current state-of-the-art for chest X-ray classification~\citep{xiaoDelvingMaskedAutoencoders2023}, a DenseNet-121~\citep{huangDenselyConnectedConvolutional2017} pre-trained on the ImageNet data set.
To predict the 14 classes, we replaced the last layer with a 14-dimensional fully-connected layer.
Before model training, images were resized if necessary.
For our experiments, we investigated the resolutions $64\times64$, $128\times128$, $256\times256$, $512\times512$, and the highest available resolution, $1024\times1024$ pixels.
We trained every model with binary cross entropy loss, AdamW~\citep{loshchilovDecoupledWeightDecay2018} optimization with default parameters, a learning rate of 0.0003, and a weight decay of 0.0001.
We divided the learning rate by a factor of ten if the validation loss did not improve after two epochs and stopped the training if the validation loss did not improve after ten epochs.
At a resolution of $64\times64$ pixels we used a batch size of 64, for $1024\times1024$ pixels of 12, and otherwise of 16.
For each model we selected the best checkpoint based on the validation area under the receiver operating curve (AUC).
We measured the effect of image resolution on chest X-ray classification using the AUC.

\subsection{Object Detection}
\begin{figure}
  \centering
  \begin{subfigure}[b]{0.48\textwidth}
    \centering
    \includegraphics[width=\linewidth]{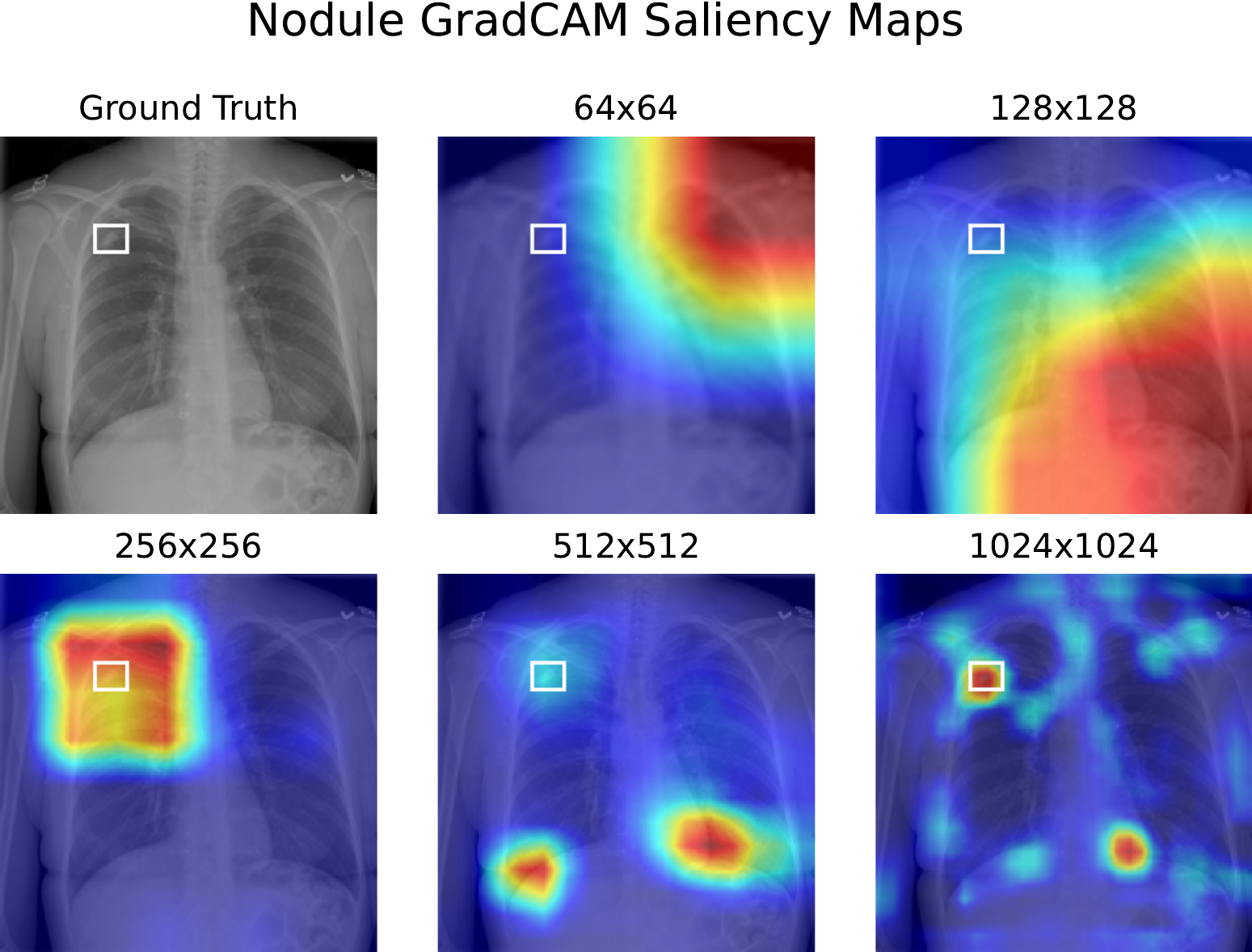}
    \caption{GradCAM saliency maps for nodule classification (white bounding box) trained with different training resolutions.}
    \label{fig:res:sal}
  \end{subfigure}\hfill
  \begin{subfigure}[b]{0.48\textwidth}
    \centering
    \includegraphics[width=\linewidth]{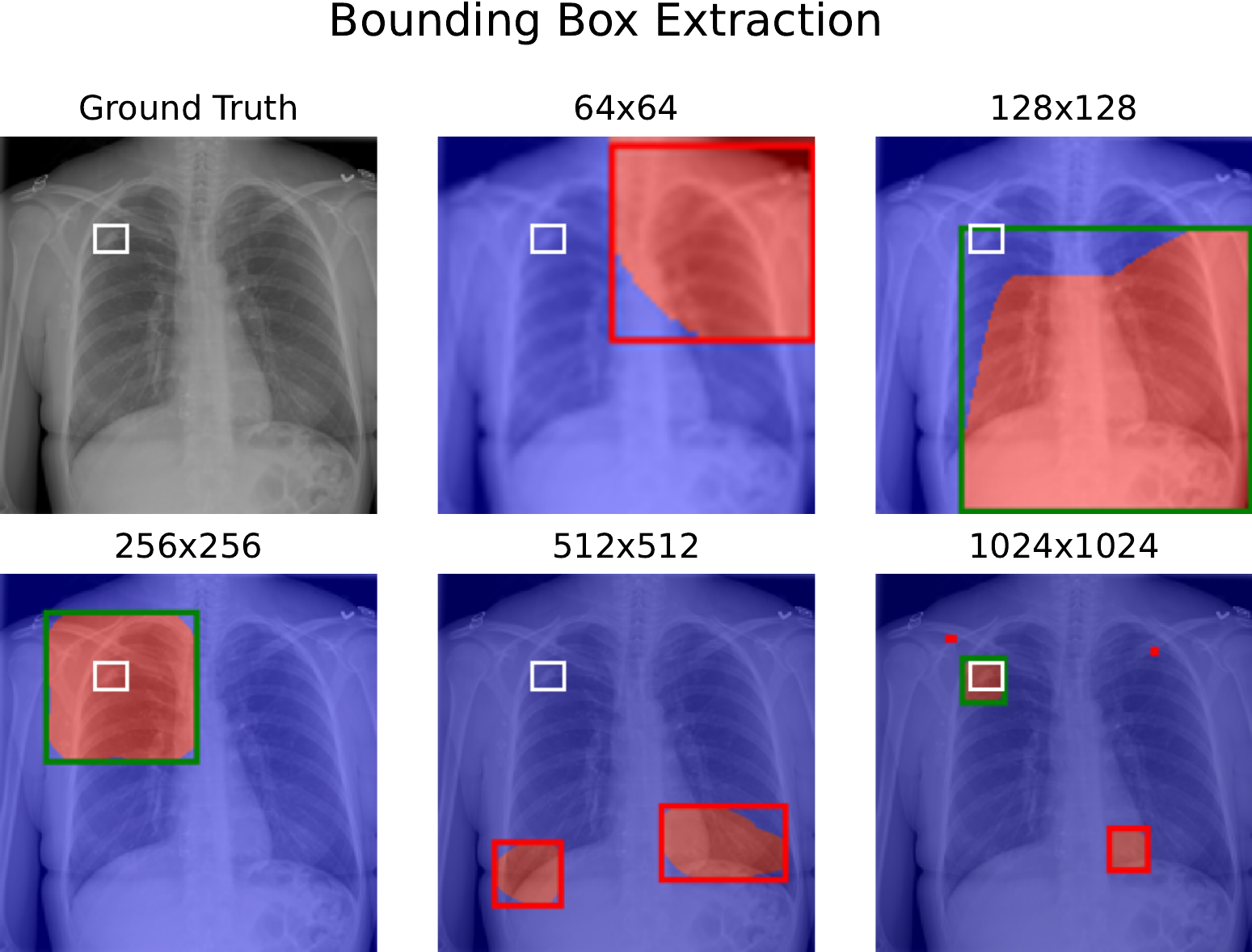}
    \caption{Saliency map-generated bounding boxes for nodule classification. Ground truth-overlapping detections surrounded in green, otherwise in red.}
    \label{fig:res:bbox}
  \end{subfigure}
  \caption{Saliency map-extracted bounding boxes for measuring the effect of image resolution on the model's basis of decision making.}
\end{figure}

Given the bounding box annotations for eight of the 14 classes, we investigated the effect of image resolution on predicted bounding boxes by comparing extracted saliency maps to the bounding box annotations.
We created segmentations from GradCAM \citep{selvarajuGradCAMVisualExplanations2017} saliency maps (see Figure~\ref{fig:res:sal}).
Saliency maps were extracted from the penultimate layer, normalized, and converted to a binary representation by applying a threshold of 0.5.
For bounding box creation, we extracted the connected components and calculated the surrounding bounding boxes, as shown in Figure~\ref{fig:res:bbox}.
For each class we measured the mean precision and accuracy.
The precision is defined as the number of true positives divided by the number of detections \citep{padillaSurveyPerformanceMetrics2020}.
Since the chest X-ray 14 data set contains only one bounding box per sample, the precision is either 0 or 1 over the number of detections.
A detection was considered positive if the intersection over union (IoU) was at least 0.1.
The IoU is defined as
\[\text{IoU}(A,B) = \frac{A\cap B}{A\cup B}, \]
where $A$ and $B$ are two bounding boxes.
Out of multiple sufficiently overlapping detections only one was considered as a true positive.

\section{Results}
\subsection{Chest X-ray Classification}
\begin{table}
  \centering
\begin{tabular}{l|rrrrr}
\backslashbox{Finding}{Resolution} &  $64\times64$ &  $128\times128$ &  $256\times256$ &  $512\times512$ &  $1024\times1024$ \\
\midrule
Atelectasis  &         0.760 &     0.800 &     0.810 &     0.807 &       \textbf{0.821} \\
Cardiomegaly &         0.858 &     0.900 &     0.906 &\textbf{0.909} &       0.908 \\
Consolidation&         0.752 &     0.787 &\textbf{0.797} & 0.794 &       \textbf{0.797} \\
Edema        &         0.866 &     0.869 &     0.885 &     0.878 &       \textbf{0.891} \\
Effusion     &         0.845 &     0.873 &     0.877 &     0.874 &       \textbf{0.879} \\
Emphysema    &         0.824 &     0.884 &     0.900 &     0.913 &       \textbf{0.937} \\
Fibrosis     &         0.748 &     0.803 &     0.816 &     0.821 &       \textbf{0.850} \\
Hernia       &         0.865 &\textbf{0.926} & 0.903 &     0.895 &       0.916 \\
Infiltration &         0.678 &     0.707 &\textbf{0.714} & 0.699 &       \textbf{0.714} \\
Mass         &         0.765 &     0.827 &\textbf{0.830} & 0.813 &       0.829 \\
Nodule       &         0.669 &     0.719 &     0.761 &     0.780 &       \textbf{0.803} \\
Pleural Thickening &   0.730 &     0.751 &     0.757 &     0.763 &       \textbf{0.796} \\
Pneumonia    &         0.688 &     0.743 &     0.760 &     0.760 &       \textbf{0.769} \\
Pneumothorax &         0.799 &     0.839 &     0.858 &     0.859 &       \textbf{0.877} \\
\midrule
Mean         &         0.775 &     0.816 &     0.827 &     0.826 &       \textbf{0.842} \\
\end{tabular}
\caption{Chest X-ray classification AUCs for different image resolutions.
  Highest values are highlighted in bold.
}
\label{tab:res:auc}
\end{table}
Per-class chest X-ray classification AUC scores are provided in Table~\ref{tab:res:auc}.
Unsurprisingly, the model trained on only $64\times64$ pixel images scored the lowest, with a mean AUC of 77.5 \%.
The highest resolution, $1024\times1024$ pixels, performed best with a mean AUC of 84.2 \%, followed by $256\times256$ pixels with a mean AUC of 82.7 \%.

\subsection{Object Detection}
\begin{figure}
  \centering
  \begin{subfigure}[b]{0.48\textwidth}
    \includegraphics[width=\textwidth]{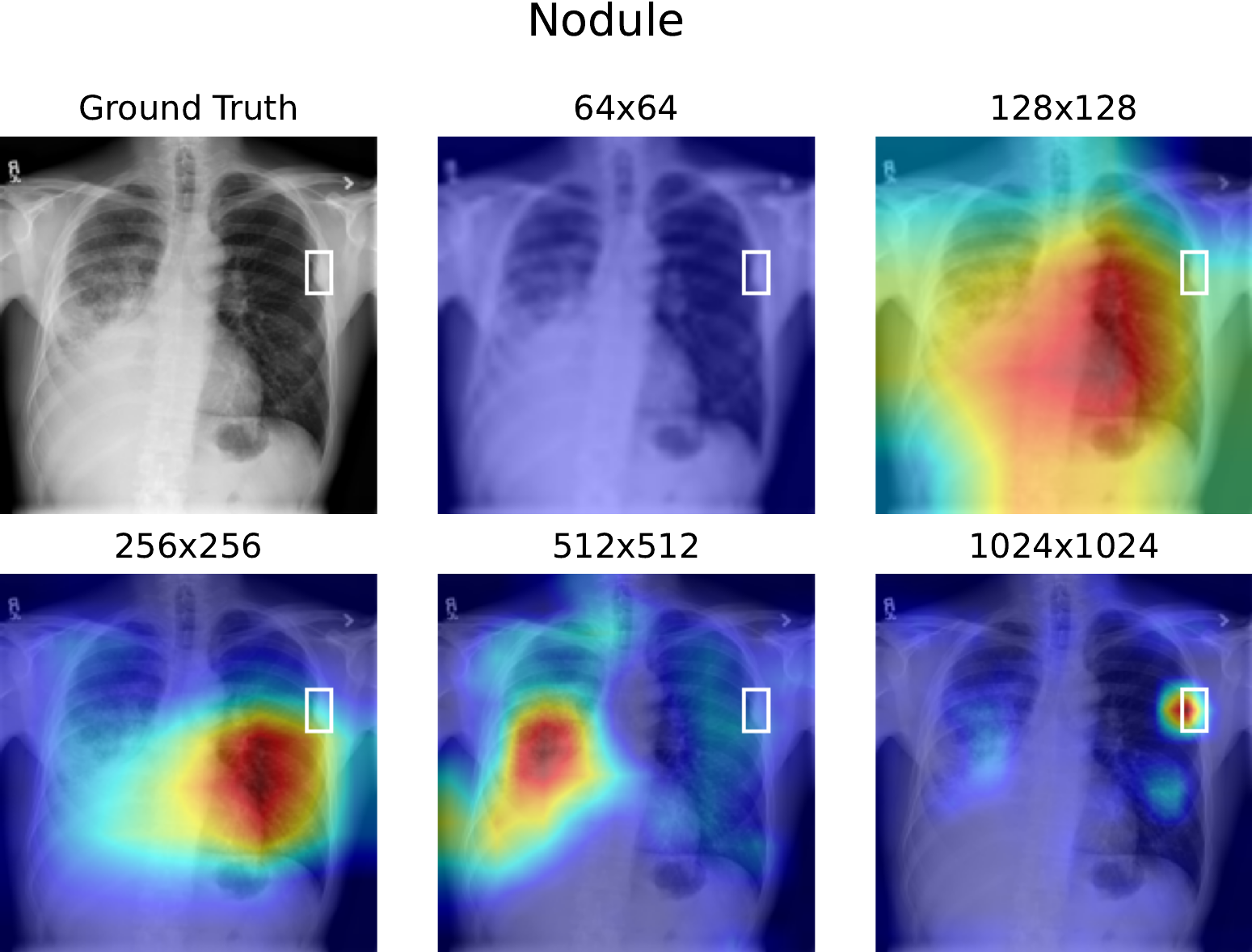}
  \end{subfigure}
  \hfill
  \begin{subfigure}[b]{0.48\textwidth}
    \includegraphics[width=\textwidth]{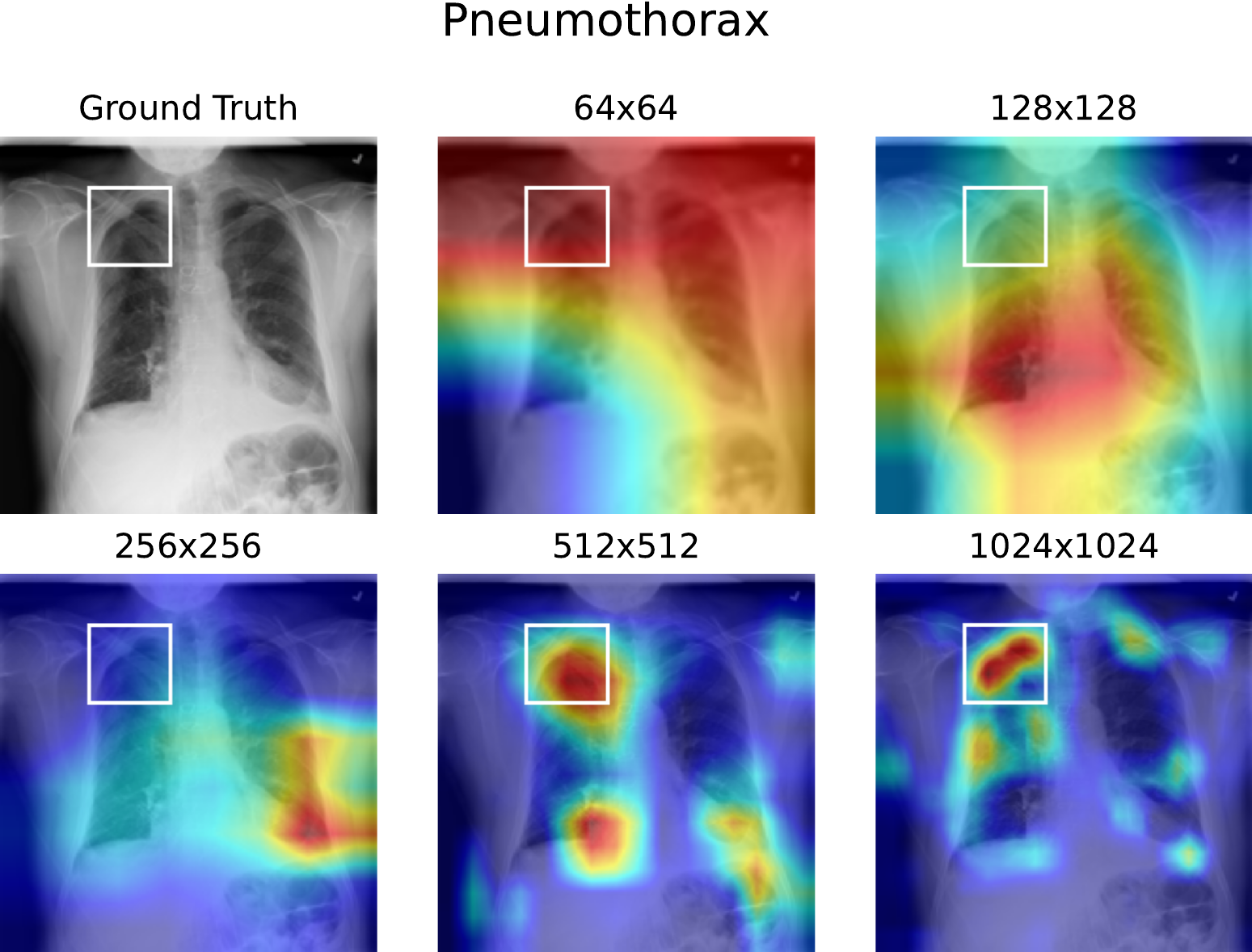}
  \end{subfigure}

  \vspace{0.5cm}

  \begin{subfigure}[b]{0.48\textwidth}
    \includegraphics[width=\textwidth]{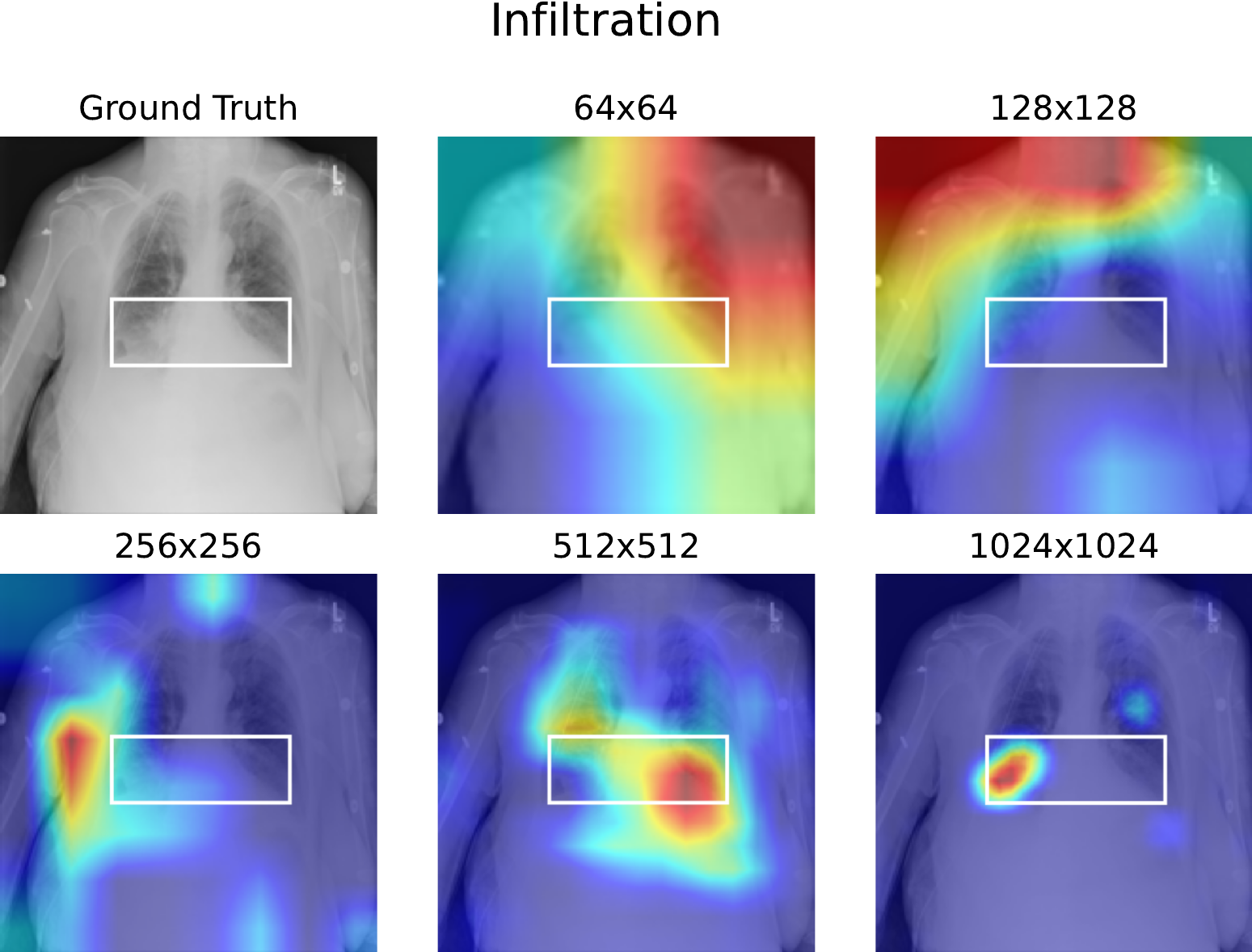}
  \end{subfigure}
  \hfill
  \begin{subfigure}[b]{0.48\textwidth}
    \includegraphics[width=\textwidth]{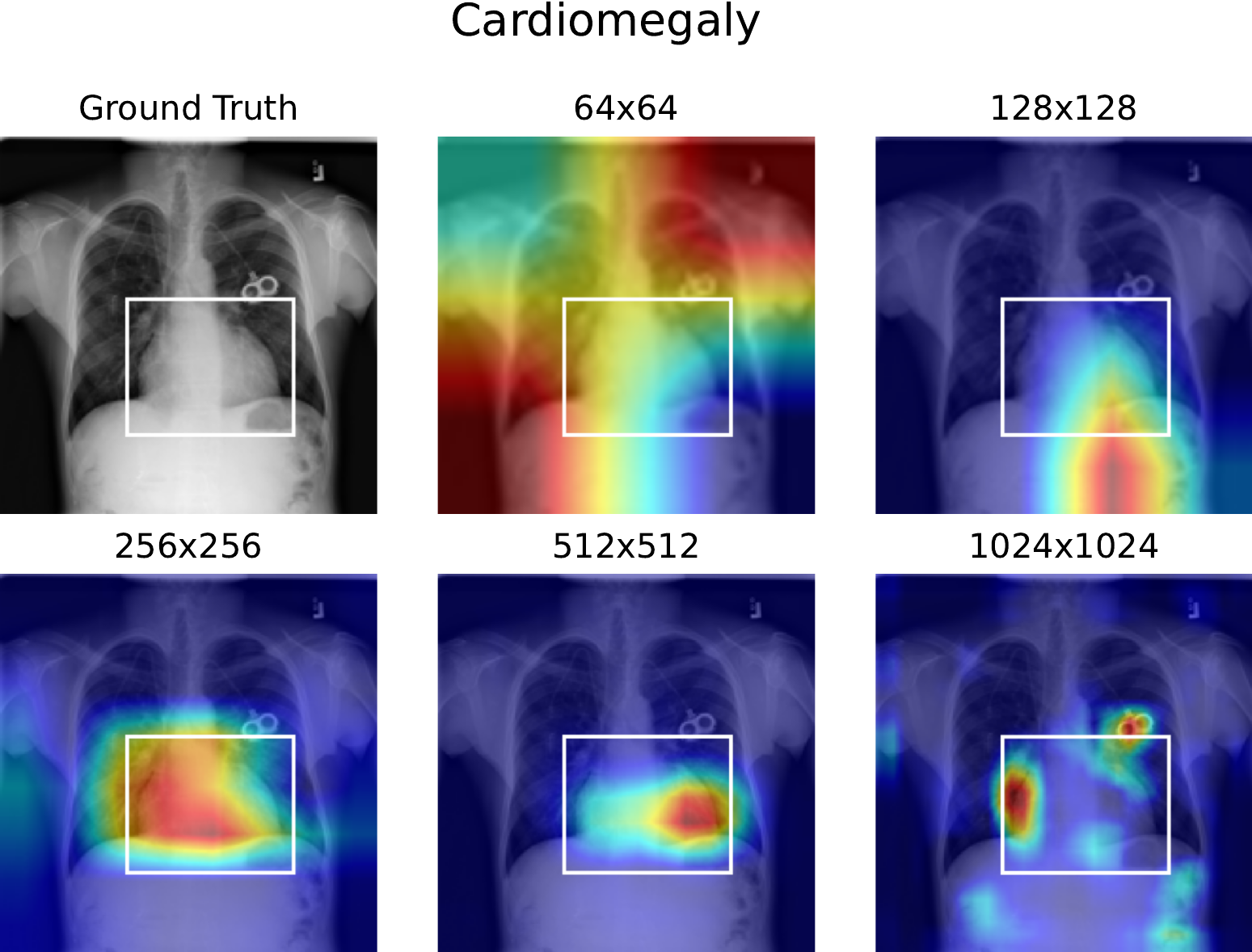}
  \end{subfigure}
  \caption{Generated saliency maps when trained with different resolutions for nodule (a), pneumothorax (b), infiltration (c), and cardiomegaly (d) classification (ground truth marked in the white bounding box).}
  \label{fig:res:saliency_example}
\end{figure}

Examples of the effect of image resolution on generated saliency maps are provided in Figure~\ref{fig:res:saliency_example}.
When increasing the image resolution, the generated GradCAM saliency maps become more granular due to the larger activation size.

\begin{table}
  \centering
\begin{tabular}{lrrrrr}
{Finding} &  $64\times64$ &  $128\times128$ &  $256\times256$ &  $512\times512$ &  $1024\times1024$ \\
\midrule
Nodule       &  0.000 &    0.000 &    0.000 &    0.150 &      \textbf{0.326} \\
Atelectasis  &  0.001 &    0.117 &    0.120 &    0.361 &      \textbf{0.395} \\
Mass         &  0.004 &    0.003 &    0.330 &    0.417 &      \textbf{0.508} \\
Pneumothorax &  0.003 &    0.061 &    \textbf{0.391} &    0.180 &      0.198 \\
Effusion     &  0.004 &    0.036 &    0.310 &    0.353 &      \textbf{0.356} \\
Infiltration &  0.008 &    \textbf{0.283} &    0.084 &    0.201 &      0.200 \\
Pneumonia    &  0.095 &    0.370 &    0.189 &    0.394 &      \textbf{0.402} \\
Cardiomegaly &  0.114 &    0.804 &    \textbf{0.946} &    0.792 &      0.242 \\
\midrule
Mean        &   0.029 &    0.209 &    0.296 & 0.356 & 0.328  \\
\end{tabular}
\caption{Mean precision at intersection over union $\ge$ 10 \% of chest pathology detections.
  Findings ordered by average ground truth bounding box size.
}
\label{tab:iou}
\end{table}

Mean precision @ IoU $\ge$ 0.1 results for pathology detection are reported in Table~\ref{tab:iou}, ordered by average ground truth bounding box size.
For 5 of 8 classes the highest resolution, $1024\times1024$ pixels, scored the highest precision, except for cardiomegaly, infiltration, and pneumothorax, where $256\times256$ (cardiomegaly, pneumothorax) and $128\times128$ (infiltration) performed best.
For the three smallest pathologies, nodule, atelectasis, and mass, the highest resolution strongly outperformed the others.
Especially for the smallest pathology, nodule, the mean precision was non-zero only for the resolutions $512\times512$ and $1024\times1024$ pixels.
Noticeably, the largest pathology, cardiomegaly, was best detected according to the saliency maps at a resolution of $256\times256$ and second to worst at $1024\times1024$ pixels.
However, training $1024\times1024$ pixel resolutions achieved a classification AUC of 90.8 \% compared to 90.6 \% when training with $256\times256$ pixel images, see Table~\ref{tab:res:auc}.

\begin{table}
  \centering
\begin{tabular}{lrrrrr}
{Finding} &  64x64 &  128x128 &  256x256 &  512x512 &  1024x1024 \\
\midrule
Nodule       &  0.000 &    0.000 &    0.000 &    0.278 &      \textbf{0.611} \\
Atelectasis  &  0.070 &    0.140 &    0.256 &    \textbf{0.605} &      0.488 \\
Mass         &  0.100 &    0.150 &    0.500 &    0.550 &      \textbf{0.600} \\
Pneumothorax &  0.167 &    0.222 &    0.500 &    0.389 &      \textbf{0.556} \\
Effusion     &  0.233 &    0.200 &    0.600 &    \textbf{0.833} &      0.567 \\
Infiltration &  0.440 &    \textbf{0.520} &    0.360 &    0.440 &      0.200 \\
Pneumonia    &  0.227 &    0.455 &    \textbf{0.727} &    0.591 &      0.591 \\
Cardiomegaly &  0.643 &    0.857 &    \textbf{0.964} &    \textbf{0.964} &      0.536 \\
\end{tabular}
\caption{Bounding box detection accuracy at intersection over union greater or equal to 0.1. Rows ordered by average bounding box size.}
\label{tab:acciou}
\end{table}

Mean detection accuracy results, ignoring multiple detections, are reported in Table~\ref{tab:acciou}.
Similarly to the mean precision results, higher resolutions ($512\times512$, $1024\times1024$) achieved a higher accuracy for smaller bounding boxes, and lower resolutions ($128\times128$, $256\times256$) for larger bounding boxes.
On average, the highest accuracy was measured when trained with $512\times512$ pixel images.

\section{Discussion}
We investigated the importance of chest X-ray resolution on classification performance motivated by small pathologies that become indistinguishable on low resolution images (see Figure~\ref{fig:res:motivation}).

The classification results show that overall a higher resolution improves image classification performance (see Table~\ref{tab:res:auc}).
On average, the highest available resolution, $1024\times1024$, performed best with an AUC of 84.2\% compared to a common resolution of $256\times256$ pixels with an AUC of 82.7 \%.
While \cite{sabottkeEffectImageResolution2020} achieved maximum AUCs between $256\times256$ and $448\times448$ pixels resolution, they tested only up to $600\times600$ pixels.
We also observed a slight decline in AUC from $256\times256$ to $512\times512$ pixel resolution for most (9/14) classes.
These findings are in line with their conclusion that a resolution $600\times600$ was not optimal.
However, our results show that an even higher image resolution, $1024\times1024$ pixels, improved chest X-ray classification performance.
Similar results were obtained for image classification accuracy on ImageNet~\citep{tanEfficientNetRethinkingModel2020}.

Surprisingly, even a resolution as low as $64\times64$ pixels achieved an classification AUC of 77.5 \%.
While one could argue that this result suggests that training at a lower resolution is a sensible performance trade-off for faster training, inspecting the saliency maps draws a different picture.
For example, the mean precision and accuracy detection results for the smallest pathology, nodule, surpass zero only at a resolution of $512\times512$ or higher (see Tables~\ref{tab:iou} and \ref{tab:acciou}).
On the one side, this is due to the intersection over union threshold that penalizes very large bounding boxes (see Figure \ref{fig:res:bbox}).
On the other side, both quantitative results and visual inspection showed that the models attend to incorrect places for the prediction at a lower resolution (see examples in Figure~\ref{fig:res:saliency_example}).
We interpret these results as that the model is forced to learn spurious distinctive features if the resolution is not sufficient.
While the detection performance decreased for larger bounding boxes when trained at a higher resolution, these models ($512\times512$ and $1024\times1024$) still achieved the highest classification AUCs.
Inspecting the saliency maps revealed that, for example, for cardiomegaly, the models attended to the correct regions but focused only on fractions of the area surrounded by the ground truth bounding boxes.
Considering both classification and detection results show that training at a higher resolution, for example $1024\times1024$ is preferable.

Our study has several limitations.
First, we measured the importance of resolution on basis of decision making by comparing saliency maps to bounding boxes.
Given more bounding box annotated data and encouraged by our results, a future study will investigate the effect of image resolution on detection performance.
Second, our experimental setup tested only resolutions up to the highest available resolution of $1024\times1024$ pixels.

In conclusion, we investigated the effect of image resolution on chest X-ray classification.
We showed that training at a higher resolution than conventionally, $1024\times1024$ pixels, achieves a higher classification performance. Furthermore, our results suggest that training at a lower but common resolution of $256\times256$ pixels poses the risk of encouraging the model to base its prediction on spurious features.

\section{Acknowledgments}
This work was supported in part by the German federal ministry of health’s program for digital innovations for the improvement of patient-centered care in healthcare [grant agreement no. 2520DAT920].

\bibliographystyle{elsarticle-harv}

\begin{thebibliography}{14}
\expandafter\ifx\csname natexlab\endcsname\relax\def\natexlab#1{#1}\fi
\providecommand{\url}[1]{\texttt{#1}}
\providecommand{\href}[2]{#2}
\providecommand{\path}[1]{#1}
\providecommand{\DOIprefix}{doi:}
\providecommand{\ArXivprefix}{arXiv:}
\providecommand{\URLprefix}{URL: }
\providecommand{\Pubmedprefix}{pmid:}
\providecommand{\doi}[1]{\href{http://dx.doi.org/#1}{\path{#1}}}
\providecommand{\Pubmed}[1]{\href{pmid:#1}{\path{#1}}}
\providecommand{\bibinfo}[2]{#2}
\ifx\xfnm\relax \def\xfnm[#1]{\unskip,\space#1}\fi
\bibitem[{Deng et~al.(2009)Deng, Dong, Socher, Li, Li and {Fei-Fei}}]{ImageNet}
\bibinfo{author}{Deng, J.}, \bibinfo{author}{Dong, W.},
  \bibinfo{author}{Socher, R.}, \bibinfo{author}{Li, L.J.},
  \bibinfo{author}{Li, K.}, \bibinfo{author}{{Fei-Fei}, L.},
  \bibinfo{year}{2009}.
\newblock \bibinfo{title}{{{ImageNet}}: {{A}} large-scale hierarchical image
  database}, in: \bibinfo{booktitle}{2009 {{IEEE Conference}} on {{Computer
  Vision}} and {{Pattern Recognition}}}, pp. \bibinfo{pages}{248--255}.
\newblock \DOIprefix\doi{10.1109/CVPR.2009.5206848}.
\bibitem[{He et~al.(2016)He, Zhang, Ren and Sun}]{heDeepResidualLearning2016}
\bibinfo{author}{He, K.}, \bibinfo{author}{Zhang, X.}, \bibinfo{author}{Ren,
  S.}, \bibinfo{author}{Sun, J.}, \bibinfo{year}{2016}.
\newblock \bibinfo{title}{Deep residual learning for image recognition}, in:
  \bibinfo{booktitle}{Proceedings of the {{IEEE}} Conference on Computer Vision
  and Pattern Recognition}, pp. \bibinfo{pages}{770--778}.
\bibitem[{Huang et~al.(2017)Huang, Liu, Van Der~Maaten and
  Weinberger}]{huangDenselyConnectedConvolutional2017}
\bibinfo{author}{Huang, G.}, \bibinfo{author}{Liu, Z.}, \bibinfo{author}{Van
  Der~Maaten, L.}, \bibinfo{author}{Weinberger, K.Q.}, \bibinfo{year}{2017}.
\newblock \bibinfo{title}{Densely connected convolutional networks}, in:
  \bibinfo{booktitle}{Proceedings of the {{IEEE}} Conference on Computer Vision
  and Pattern Recognition}, pp. \bibinfo{pages}{4700--4708}.
\newblock \href{http://arxiv.org/abs/1608.06993}{{\tt arXiv:1608.06993}}.
\bibitem[{Johnson et~al.(2019)Johnson, Pollard, Berkowitz, Greenbaum, Lungren,
  Deng, Mark and Horng}]{johnsonMIMICCXRDeidentifiedPublicly2019}
\bibinfo{author}{Johnson, A.E.}, \bibinfo{author}{Pollard, T.J.},
  \bibinfo{author}{Berkowitz, S.J.}, \bibinfo{author}{Greenbaum, N.R.},
  \bibinfo{author}{Lungren, M.P.}, \bibinfo{author}{Deng, C.y.},
  \bibinfo{author}{Mark, R.G.}, \bibinfo{author}{Horng, S.},
  \bibinfo{year}{2019}.
\newblock \bibinfo{title}{{{MIMIC-CXR}}, a de-identified publicly available
  database of chest radiographs with free-text reports}.
\newblock \bibinfo{journal}{Scientific Data} \bibinfo{volume}{6}.
\bibitem[{Krizhevsky et~al.(2012)Krizhevsky, Sutskever and
  Hinton}]{krizhevskyImagenetClassificationDeep2012}
\bibinfo{author}{Krizhevsky, A.}, \bibinfo{author}{Sutskever, I.},
  \bibinfo{author}{Hinton, G.E.}, \bibinfo{year}{2012}.
\newblock \bibinfo{title}{Imagenet classification with deep convolutional
  neural networks}.
\newblock \bibinfo{journal}{Advances in neural information processing systems}
  \bibinfo{volume}{25}, \bibinfo{pages}{1097--1105}.
\bibitem[{Loshchilov and Hutter(2018)}]{loshchilovDecoupledWeightDecay2018}
\bibinfo{author}{Loshchilov, I.}, \bibinfo{author}{Hutter, F.},
  \bibinfo{year}{2018}.
\newblock \bibinfo{title}{Decoupled {{Weight Decay Regularization}}}, in:
  \bibinfo{booktitle}{International {{Conference}} on {{Learning
  Representations}}}.
\bibitem[{Padilla et~al.(2020)Padilla, Netto and {da
  Silva}}]{padillaSurveyPerformanceMetrics2020}
\bibinfo{author}{Padilla, R.}, \bibinfo{author}{Netto, S.L.},
  \bibinfo{author}{{da Silva}, E.A.}, \bibinfo{year}{2020}.
\newblock \bibinfo{title}{A survey on performance metrics for object-detection
  algorithms}, in: \bibinfo{booktitle}{2020 {{International Conference}} on
  {{Systems}}, {{Signals}} and {{Image Processing}} ({{IWSSIP}})},
  \bibinfo{publisher}{{IEEE}}. pp. \bibinfo{pages}{237--242}.
\bibitem[{Sabottke and Spieler(2020a)}]{sabottkeEffectImageResolution2020a}
\bibinfo{author}{Sabottke, C.F.}, \bibinfo{author}{Spieler, B.M.},
  \bibinfo{year}{2020}a.
\newblock \bibinfo{title}{The {{Effect}} of {{Image Resolution}} on {{Deep
  Learning}} in {{Radiography}}}.
\newblock \bibinfo{journal}{Radiology: Artificial Intelligence}
  \bibinfo{volume}{2}, \bibinfo{pages}{e190015}.
\newblock \DOIprefix\doi{10.1148/ryai.2019190015}.
\bibitem[{Sabottke and Spieler(2020b)}]{sabottkeEffectImageResolution2020}
\bibinfo{author}{Sabottke, C.F.}, \bibinfo{author}{Spieler, B.M.},
  \bibinfo{year}{2020}b.
\newblock \bibinfo{title}{The {{Effect}} of {{Image Resolution}} on {{Deep
  Learning}} in {{Radiography}}}.
\newblock \bibinfo{journal}{Radiology: Artificial Intelligence}
  \bibinfo{volume}{2}, \bibinfo{pages}{e190015}.
\newblock \DOIprefix\doi{10.1148/ryai.2019190015}.
\bibitem[{Selvaraju et~al.(2017)Selvaraju, Cogswell, Das, Vedantam, Parikh and
  Batra}]{selvarajuGradCAMVisualExplanations2017}
\bibinfo{author}{Selvaraju, R.R.}, \bibinfo{author}{Cogswell, M.},
  \bibinfo{author}{Das, A.}, \bibinfo{author}{Vedantam, R.},
  \bibinfo{author}{Parikh, D.}, \bibinfo{author}{Batra, D.},
  \bibinfo{year}{2017}.
\newblock \bibinfo{title}{Grad-{{CAM}}: {{Visual Explanations From Deep
  Networks}} via {{Gradient-Based Localization}}}, in:
  \bibinfo{booktitle}{Proceedings of the {{IEEE International Conference}} on
  {{Computer Vision}}}, pp. \bibinfo{pages}{618--626}.
\bibitem[{Tan and Le(2020)}]{tanEfficientNetRethinkingModel2020}
\bibinfo{author}{Tan, M.}, \bibinfo{author}{Le, Q.V.}, \bibinfo{year}{2020}.
\newblock \bibinfo{title}{{{EfficientNet}}: {{Rethinking Model Scaling}} for
  {{Convolutional Neural Networks}}}.
\newblock \bibinfo{journal}{arXiv:1905.11946 [cs, stat]}
  \href{http://arxiv.org/abs/1905.11946}{{\tt arXiv:1905.11946}}.
\bibitem[{Wang et~al.(2017)Wang, Peng, Lu, Lu, Bagheri and
  Summers}]{wangChestXray8HospitalscaleChest2017}
\bibinfo{author}{Wang, X.}, \bibinfo{author}{Peng, Y.}, \bibinfo{author}{Lu,
  L.}, \bibinfo{author}{Lu, Z.}, \bibinfo{author}{Bagheri, M.},
  \bibinfo{author}{Summers, R.M.}, \bibinfo{year}{2017}.
\newblock \bibinfo{title}{{{ChestX-ray8}}: {{Hospital-scale Chest X-ray
  Database}} and {{Benchmarks}} on {{Weakly-Supervised Classification}} and
  {{Localization}} of {{Common Thorax Diseases}}}, in: \bibinfo{booktitle}{2017
  {{IEEE Conference}} on {{Computer Vision}} and {{Pattern Recognition}}
  ({{CVPR}})}, pp. \bibinfo{pages}{3462--3471}.
\newblock \DOIprefix\doi{10.1109/CVPR.2017.369},
  \href{http://arxiv.org/abs/1705.02315}{{\tt arXiv:1705.02315}}.
\bibitem[{Wollek et~al.(2023)Wollek, Graf, {\v C}e{\v c}atka, Fink, Willem,
  Sabel and Lasser}]{wollekAttentionbasedSaliencyMaps2023}
\bibinfo{author}{Wollek, A.}, \bibinfo{author}{Graf, R.}, \bibinfo{author}{{\v
  C}e{\v c}atka, S.}, \bibinfo{author}{Fink, N.}, \bibinfo{author}{Willem, T.},
  \bibinfo{author}{Sabel, B.O.}, \bibinfo{author}{Lasser, T.},
  \bibinfo{year}{2023}.
\newblock \bibinfo{title}{Attention-based {{Saliency Maps Improve
  Interpretability}} of {{Pneumothorax Classification}}}.
\newblock \bibinfo{journal}{Radiology: Artificial Intelligence} ,
  \bibinfo{pages}{e220187}\DOIprefix\doi{10.1148/ryai.220187}.
\bibitem[{Xiao et~al.(2023)Xiao, Bai, Yuille and
  Zhou}]{xiaoDelvingMaskedAutoencoders2023}
\bibinfo{author}{Xiao, J.}, \bibinfo{author}{Bai, Y.}, \bibinfo{author}{Yuille,
  A.}, \bibinfo{author}{Zhou, Z.}, \bibinfo{year}{2023}.
\newblock \bibinfo{title}{Delving into masked autoencoders for multi-label
  thorax disease classification}, in: \bibinfo{booktitle}{Proceedings of the
  {{IEEE}}/{{CVF Winter Conference}} on {{Applications}} of {{Computer
  Vision}}}, pp. \bibinfo{pages}{3588--3600}.

\end{thebibliography}

\end{document}